\documentclass{article}

\usepackage{graphicx}
\usepackage{amsmath}
\usepackage{pgfplots}
\pgfplotsset{compat=1.18}
\usepackage{amssymb}
\usepackage{xcolor}

 \usepackage[preprint]{neurips_2026}

\usepackage[utf8]{inputenc} 
\usepackage[T1]{fontenc}    
\usepackage{hyperref} 
\usepackage{url}            
\usepackage{booktabs}       
\usepackage{tabularx}
\newcolumntype{P}[1]{>{\raggedright\arraybackslash}p{#1}}
\usepackage{amsfonts}       
\usepackage{nicefrac}       
\usepackage{microtype}      
\usepackage{xcolor}         

\title{The Detectability Gap: Hidden Heterogeneity in Hallucination Detection Across Language Models}

\workshoptitle{GlobalSouthAI @ NeurIPS 2026: Rethinking AI for and from the Global South}

\author{%
  \textbf{Pranav Darshan}$^{1}$ \quad \textbf{Pranav A}$^{1}$ \quad \textbf{Sravan Karthick T}$^{1}$ \\
  \textbf{Minal Moharir}$^{1}$ \quad \textbf{Ivan P.~Yamshchikov}$^{2}$ \\[4pt]
  $^{1}$Department of Computer Science and Engineering,
  R.V.\ College of Engineering, India \\
  $^{2}$CAIRO, Technical University of Applied Sciences
  W\"urzburg-Schweinfurt, Germany \\[3pt]
  \texttt{\{pranavdarshan.cs22, pranava.cs21, sravankt.cs20, minalmoharir\}@rvce.edu.in} \\
  \texttt{ivan.yamshchikov@thws.de}
}

\begin{document}
\raggedbottom

\maketitle

\begin{abstract}

Sampling based consistency is widely used for hallucination detection, yet aggregate performance can conceal systematic differences in which errors are detectable. This work studies that heterogeneity across four language models and three factual question answering datasets. Partitioning hallucinations by answer agreement reveals high agreement (\textbf{Ghost}) and low agreement (\textbf{Flickering}) regimes with an apparent detectability gap of $0.35$ to $0.46$ AUC. Because the statistics used to define the regimes and measure this gap are strongly coupled ($|\rho|\approx0.94$ to $1.00$), the raw result is treated as a property of agreement based detection rather than independent evidence. After freezing regime assignments, lexical and semantic response dispersion preserve the asymmetry, with bootstrap $95\%$ intervals excluding zero in all $12$ model and dataset settings. A stricter test using individual diffusion trajectories and no cross seed information preserves the asymmetry across all three LLaDA datasets ($p<0.005$) and directionally across all three Dream datasets, with one reaching significance. The hard regime varies substantially in prevalence across models ($16\%$ to $77\%$), and matched prompts frequently change regimes between models. These findings show that aggregate detection metrics conceal persistent, model dependent heterogeneity in language model failures and motivate regime conditioned evaluation.

\end{abstract}

\section{Introduction}

\label{sec}

Hallucination detectors often use \emph{stochastic consistency}: sample a model several times and flag answers that fail to reproduce
\citep{manakul2023selfcheckgpt, wang2023selfconsistency, farquhar2024detecting, zhang2023sac3}. A related line of work detects hallucinations from the denoising trace of diffusion language models directly \citep{chang2025tracedet, hemmat2026tdgnet, qian2026dynhd}. These works establish that consistency signals can miss confidently repeated errors and that diffusion trajectories carry detection relevant information; the present study is complementary; it shows that such errors form a distinct, model dependent behavioral population whose detectability is quantitatively lower for both a cross seed agreement signal and a within seed trajectory signal, and it measures how large that gap is and how it survives several independent operationalizations. Aggregate scores can conceal systematic differences in which errors are detectable. We examine whether stochastic generations reveal distinct hallucination populations, and whether their detectability differs when evaluation is separated from the information defining those populations.

Our motivation is to support the development and deployment of trustworthy and reliable AI systems globally. Uneven infrastructure, skills, and governance capacity can amplify AI risks~\citep{undp2025divergence}. Where independent verification is difficult, confidently repeated errors may invite misplaced trust. The goal is to inform affordable, effective safeguards that flag hallucinations before users rely on them. Rather than assuming uniformly higher trust or lower AI literacy across these diverse communities, we study a technical obstacle: errors that evade otherwise useful detection signals.

We evaluate four instruction tuned diffusion and autoregressive models on TriviaQA, HotpotQA, and PopQA. Generations differ only in sampling seed. Hallucinations are grouped into high agreement (\emph{Ghost}) and low agreement (\emph{Flickering}) regimes. The primary contributions are:
\begin{enumerate}
\item We expose persistent detectability differences between Ghost and Flickering hallucinations, motivating regime specific evaluation rather than aggregate scores alone.
\item We examine the circularity of raw agreement and test the asymmetry with two decoupled response level measures across all $12$ settings, then with individual diffusion trajectories without cross seed information.
\item We show that regime prevalence and the regime assigned to the same prompt depend on the model.
\end{enumerate}

\section{Experimental Setup}
\label{sec:setup}

\paragraph{Models and datasets.}
The study evaluates four instruction tuned language models: two masked
diffusion models, LLaDA 8B~\citep{nie2025llada} and Dream 7B~\citep{ye2025dream},
and two autoregressive models, Qwen2.5 7B~\citep{qwen2025qwen25} and Llama
3.1 8B~\citep{grattafiori2024llama}. Evaluation is conducted on three open
domain factual question answering datasets: TriviaQA~\citep{joshi2017triviaqa},
HotpotQA~\citep{yang2018hotpotqa}, and PopQA~\citep{mallen2023popqa}. For each
prompt, $K=3$ generations are sampled with different random seeds while all
other decoding conditions are held fixed. The autoregressive models provide
agreement level controls, while only the diffusion models are used for the
trajectory analysis.

\paragraph{Correctness and hallucination labels.}
A generation is considered correct when its answer contains a gold alias. A
prompt is labeled as a hallucination ($Y=1$) when none of its sampled
generations contains the gold answer. Alias matching is performed at the token
level for the diffusion corpora and using a case insensitive substring match
for the autoregressive corpora. Gold answers are used only for labeling and
evaluation, never as input to a detection signal.

\paragraph{Behavioral regimes.}
For each generation, a gold free answer span is extracted and compared across
seeds using token level Jaccard similarity at threshold $\tau$. Let
$A_1,\ldots,A_K$ denote the answer spans and let $\mathcal{C}_\tau$ denote the
resulting single linkage clusters. The plurality fraction is defined as
\begin{equation}
\hat{\theta}_K
=
\frac{1}{K}\max_{c\in\mathcal{C}_\tau}|c|.
\end{equation}
Among hallucinated prompts, two operational behavioral regimes are defined:
\begin{equation}
\textbf{Ghost}: \hat{\theta}_K>\frac{1}{2},
\qquad
\textbf{Flickering}: \hat{\theta}_K\leq\frac{1}{2}.
\end{equation}
Ghost hallucinations therefore contain a dominant repeated answer, whereas
Flickering hallucinations are more dispersed across seeds. Table~\ref{tab:main-examples}
illustrates the distinction. These are operational behavioral partitions and
do not imply distinct latent mechanisms.

\begin{table}[t]
\centering
\footnotesize
\setlength{\tabcolsep}{3pt}
\renewcommand{\arraystretch}{1}
\begin{tabularx}{\columnwidth}{@{}l >{\raggedright\arraybackslash}X l l@{}}
\toprule
\textbf{Type} & \textbf{Question} & \textbf{Three sampled answers} & \textbf{Gold} \\
\midrule
Ghost &
What sport does Masahito Noto play? &
baseball / baseball / baseball &
football \\
Flickering &
Who composed ``The Mission''? &
various / James Horner / The Doors &
Ennio Morricone \\
\bottomrule
\end{tabularx}
\caption{Ghost hallucinations repeat incorrect answers; Flickering
hallucinations produce different incorrect answers. No sampled answer
contains the gold answer.}
\label{tab:main-examples}
\end{table}

\paragraph{Evaluation signals.}
For a signal $s$, detectability within regime $R$ is measured by the AUC
separating hallucinations in $R$ from correct answers. The detectability gap is
\begin{equation}
\Delta(s)
=
\mathrm{AUC}_{\mathrm{Flickering}}(s)
-
\mathrm{AUC}_{\mathrm{Ghost}}(s).
\end{equation}
The raw agreement analysis uses pairwise disagreement across sampled answers.
Because this statistic is closely coupled to the plurality fraction that
defines the regimes, it is treated as descriptive. The main analysis instead
freezes the regime assignments and evaluates lexical and semantic response
dispersion that do not reuse the regime defining computation. Bootstrap
$95\%$ confidence intervals are computed at the prompt level using $B=2000$
resamples. A separate diffusion only analysis uses denoising trajectory
features from individual seeds without information across seeds.

\section{The Detectability Gap and Its Circularity}
\label{sec:gap}

\paragraph{Agreement based detectability.}
The raw disagreement signal $1-\hat{\pi}$ separates hallucinations from
correct answers much more effectively in the Flickering regime than in the
Ghost regime. Across the evaluated model and dataset combinations, Ghost AUC
ranges from $0.34$ to $0.55$, while Flickering AUC ranges from $0.80$ to
$0.96$. The resulting detectability gap is consistently large, ranging from
$0.35$ to $0.46$ AUC, with all bootstrap $95\%$ confidence intervals excluding
zero. Thus, an aggregate agreement signal can appear effective while
performing poorly on a substantial population of hallucinations.

\paragraph{The role of circularity.}
The raw gap cannot be interpreted as independent evidence of heterogeneous
detectability because the same answer agreement structure defines the regimes
and constructs the disagreement signal. Ghost hallucinations are defined by
large $\hat{\theta}$, while the evaluated signal is its agreement based
complement, $1-\hat{\pi}$. Their correlation ranges from $-0.94$ to $-1.00$
across settings.

The raw agreement result is therefore treated as descriptive. The stronger
test freezes the Ghost and Flickering assignments and changes the evaluation
signal so that it does not reuse answer span clustering, plurality, or the
agreement threshold.

\section{Detectability Heterogeneity Beyond Agreement}
\label{sec:beyond}

The central test freezes the Ghost and Flickering assignments and replaces the
agreement signal with measurements that do not reuse the regime defining
computation.

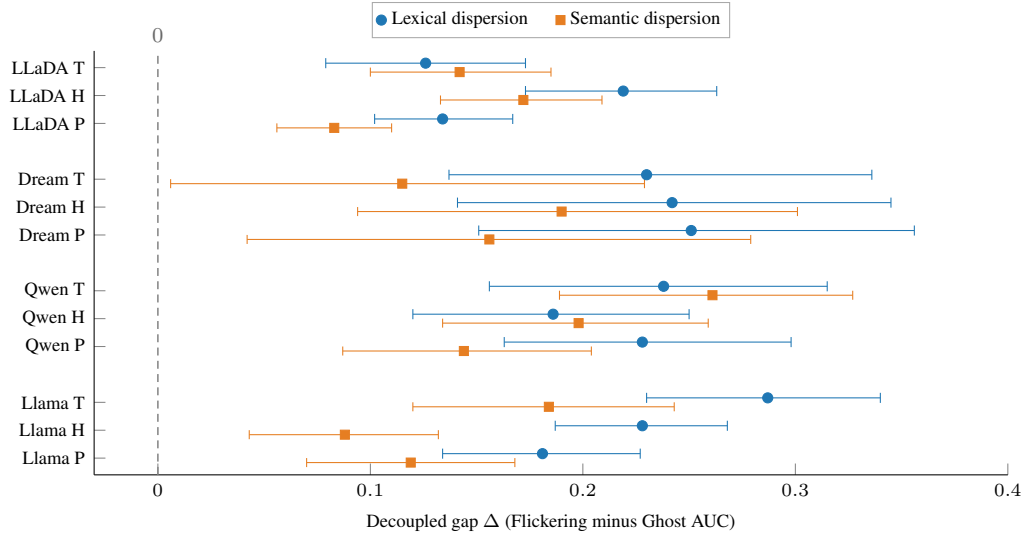
\begin{figure}[tb]
  \centering
  \resizebox{\columnwidth}{!}{

\definecolor{cBlue}{RGB}{31,119,180}
\definecolor{cOrange}{RGB}{230,126,34}
\definecolor{cInk}{RGB}{40,40,40}

\begin{tikzpicture}
\begin{axis}[
  width=0.98\linewidth,
  height=7.2cm,
  xmin=-0.03,
  xmax=0.40,
  ymin=0.4,
  ymax=15.6,
  xtick={0,0.1,0.2,0.3,0.4},
  xlabel={Decoupled gap $\Delta$ (Flickering minus Ghost AUC)},
  ytick={15,14,13,11,10,9,7,6,5,3,2,1},
  yticklabels={
    LLaDA T, LLaDA H, LLaDA P,
    Dream T, Dream H, Dream P,
    Qwen T, Qwen H, Qwen P,
    Llama T, Llama H, Llama P
  },
  tick label style={font=\scriptsize},
  label style={font=\scriptsize},
  axis y line*=left,
  axis x line*=bottom,
  x tick label style={/pgf/number format/fixed},
  clip=false,
  legend style={
    at={(0.5,1.03)},
    anchor=south,
    legend columns=2,
    draw=cInk!40,
    font=\scriptsize,
    inner sep=1.5pt,
    /tikz/every even column/.append style={column sep=3mm}
  },
]

  \draw[
    densely dashed,
    cInk!55,
    line width=0.6pt
  ]
    (axis cs:0,0.4) -- (axis cs:0,15.6);

  \node[
    font=\footnotesize,
    cInk!70,
    anchor=south
  ] at (axis cs:0,15.6) {$0$};

  \addplot[
    only marks,
    mark=*,
    mark size=1.9pt,
    cBlue,
    error bars/.cd,
      x dir=both,
      x explicit
  ]
  table[
    x=x,
    y=y,
    x error plus=ep,
    x error minus=em,
    row sep=\\
  ]{
    x     y      ep     em    \\
    0.126 15.16  0.047  0.047 \\
    0.219 14.16  0.044  0.046 \\
    0.134 13.16  0.033  0.032 \\
    0.230 11.16  0.106  0.093 \\
    0.242 10.16  0.103  0.101 \\
    0.251 9.16   0.105  0.100 \\
    0.238 7.16   0.077  0.082 \\
    0.186 6.16   0.064  0.066 \\
    0.228 5.16   0.070  0.065 \\
    0.287 3.16   0.053  0.057 \\
    0.228 2.16   0.040  0.041 \\
    0.181 1.16   0.046  0.047 \\
  };
  \addlegendentry{Lexical dispersion}

  \addplot[
    only marks,
    mark=square*,
    mark size=1.7pt,
    cOrange,
    error bars/.cd,
      x dir=both,
      x explicit
  ]
  table[
    x=x,
    y=y,
    x error plus=ep,
    x error minus=em,
    row sep=\\
  ]{
    x     y      ep     em    \\
    0.142 14.84  0.043  0.042 \\
    0.172 13.84  0.037  0.039 \\
    0.083 12.84  0.027  0.027 \\
    0.115 10.84  0.114  0.109 \\
    0.190 9.84   0.111  0.096 \\
    0.156 8.84   0.123  0.114 \\
    0.261 6.84   0.066  0.072 \\
    0.198 5.84   0.061  0.064 \\
    0.144 4.84   0.060  0.057 \\
    0.184 2.84   0.059  0.064 \\
    0.088 1.84   0.044  0.045 \\
    0.119 0.84   0.049  0.049 \\
  };
  \addlegendentry{Semantic dispersion}

\end{axis}
\end{tikzpicture}}
  \caption{Decoupled gap $\Delta$, defined as Flickering AUC minus Ghost AUC.
  Bootstrap $95\%$ confidence intervals are shown for lexical and semantic
  dispersion. All twelve model and dataset settings have positive gaps.}
  \label{fig:forest}
\end{figure}

\paragraph{Decoupled evaluation.}
Two response level measures are considered. Lexical dispersion is the mean
pairwise $1-$Jaccard over content words in the whole responses. It does not
use the answer span heuristic, cluster sizes, plurality fraction, or agreement
threshold. Semantic dispersion is the mean pairwise cosine distance between
all MiniLM L6 v2 embeddings of the whole \mbox{responses~\citep{reimers2019sbert}}.
Although both measures capture variation across stochastic generations,
neither reuses the computation that defines the regimes.

The lexical gap is $+0.13$ to $+0.29$ and the semantic gap is $+0.08$ to
$+0.26$. Both signals favor Flickering in every model and dataset combination,
and all bootstrap $95\%$ confidence intervals exclude zero. Figure~\ref{fig:forest}
summarizes the result. The asymmetry is smaller than the raw agreement gap,
but its direction does not change when the evaluation signal is changed.

\section{Within Seed Evidence}
\label{sec:trajectory}

The response level analysis shows that the detectability gap survives
measurements separated from the regime definition. Diffusion models provide a
stricter test because each generation exposes a denoising trajectory. The
trajectory signal therefore uses information from a single seed without
access to the other sampled seeds.

\paragraph{Trajectory only evaluation.}
The trajectory features capture when answer tokens become committed, confidence
before commitment, and confidence change at commitment. No information about
the other seeds, answer agreement, clustering, or plurality is provided to the
detector. A leakage free out of fold logistic detector separates hallucinations
from correct answers, after which predictions are evaluated separately for the
frozen Ghost and Flickering groups.

On LLaDA, the trajectory only gap is positive on all three datasets, with
$\Delta=0.106$, $0.123$, and $0.092$ on TriviaQA, HotpotQA, and PopQA,
respectively. All three permutation tests reach $p<0.005$. Dream shows the
same direction on all three datasets, with gaps of $0.035$, $0.072$, and
$0.163$, although only PopQA reaches significance ($p=0.003$). Complete
results and confidence intervals are reported in Table~\ref{tab:trajectory-app}.

Importantly, Ghost AUC rises to $0.60$ to $0.79$ under trajectory information.
Ghost hallucinations are therefore not inherently undetectable. They remain
harder to detect than Flickering hallucinations even when information across
seeds is removed from the signal.

\section{Model Dependence of the Hard Regime}
\label{sec:crossmodel}

The detectability asymmetry is consistent across models, but the composition
of the hallucination population is not. The proportion of Ghost hallucinations
ranges from $65$ to $77\%$ for LLaDA, $59$ to $77\%$ for Qwen, and $62$ to
$67\%$ for Llama, while Dream has only $16$ to $30\%$ Ghost hallucinations.
Thus, the model with the smallest hard regime still exhibits the same positive
decoupled detectability gaps.

The dependence on the model is also visible when the same prompts are
evaluated by different models. Among $347$ prompts hallucinated by both LLaDA
and Dream, $181$ move from Ghost under LLaDA to Flickering under Dream,
whereas only $17$ make the reverse transition. A McNemar test gives
$p<10^{-35}$. Only $67$ Ghost cases remain Ghost under both models. These
transitions show that, within this matched comparison, regime membership is
not determined solely by the question. The generation model itself plays an
important role in determining whether a hallucination is stable or variable
across seeds.

\section{Conclusion}
\label{sec:conclusion}

These results show that hallucination detectability is heterogeneous across
behavioral regimes and that regime membership itself depends on the generation
model. The raw agreement gap is strongly coupled to the regime definition, but
the asymmetry persists under lexical and semantic response level measurements
that do not reuse that computation. A stricter trajectory only analysis also
preserves the asymmetry for LLaDA and shows the same direction across all
three Dream datasets.

For accessible safeguards in the Global South and other resource constrained
settings, detection must address consistently wrong answers, not only variable
ones. Regime specific evaluation and trajectory based warnings are steps toward
this goal, rather than guarantees of safe use. Deployment affordability and
performance on locally relevant languages and tasks remain to be evaluated.

\bibliographystyle{plainnat}
\bibliography{references}

@inproceedings{nie2025llada,
  title     = {Large Language Diffusion Models},
  author    = {Nie, Shen and Zhu, Fengqi and You, Zebin and Zhang, Xiaolu and Ou, Jingyang and Hu, Jun and Zhou, Jun and Lin, Yankai and Wen, Ji Rong and Li, Chongxuan},
  booktitle = {Proceedings of the 42nd International Conference on Machine Learning (ICML)},
  year      = {2025},
  url       = {https://arxiv.org/abs/2502.09992}
}

@article{qwen2025qwen25,
  title   = {Qwen2.5 Technical Report},
  author  = {Qwen Team},
  journal = {arXiv preprint arXiv:2412.15115},
  year    = {2025},
  url     = {https://arxiv.org/abs/2412.15115}
}

@article{grattafiori2024llama,
  title   = {The Llama 3 Herd of Models},
  author  = {Grattafiori, Aaron and Dubey, Abhimanyu and Jauhri, Abhinav and others},
  journal = {arXiv preprint arXiv:2407.21783},
  year    = {2024},
  url     = {https://arxiv.org/abs/2407.21783}
}

@inproceedings{reimers2019sbert,
  title     = {Sentence {BERT}: Sentence Embeddings using Siamese {BERT} Networks},
  author    = {Reimers, Nils and Gurevych, Iryna},
  booktitle = {Proceedings of the 2019 Conference on Empirical Methods in Natural Language Processing (EMNLP)},
  year      = {2019},
  pages     = {3982--3992},
  url       = {https://arxiv.org/abs/1908.10084}
}

@article{farquhar2024detecting,
  title   = {Detecting Hallucinations in Large Language Models Using Semantic Entropy},
  author  = {Farquhar, Sebastian and Kossen, Jannik and Kuhn, Lorenz and Gal, Yarin},
  journal = {Nature},
  volume  = {630},
  pages   = {625--630},
  year    = {2024},
  doi     = {10.1038/s41586-024-07421-0}
}

@inproceedings{wang2023selfconsistency,
  title     = {Self Consistency Improves Chain of Thought Reasoning in Language Models},
  author    = {Wang, Xuezhi and Wei, Jason and Schuurmans, Dale and Le, Quoc and Chi, Ed and Narang, Sharan and Chowdhery, Aakanksha and Zhou, Denny},
  booktitle = {International Conference on Learning Representations (ICLR)},
  year      = {2023},
  url       = {https://arxiv.org/abs/2203.11171}
}

@inproceedings{joshi2017triviaqa,
  title     = {TriviaQA: A Large Scale Distantly Supervised Challenge Dataset for Reading Comprehension},
  author    = {Joshi, Mandar and Choi, Eunsol and Weld, Daniel S. and Zettlemoyer, Luke},
  booktitle = {Proceedings of the 55th Annual Meeting of the Association for Computational Linguistics (ACL)},
  year      = {2017},
  pages     = {1601--1611},
  doi       = {10.18653/v1/P17-1147}
}

@inproceedings{yang2018hotpotqa,
  title     = {{HotpotQA}: A Dataset for Diverse, Explainable Multi hop Question Answering},
  author    = {Yang, Zhilin and Qi, Peng and Zhang, Saizheng and Bengio, Yoshua and Cohen, William W. and Salakhutdinov, Ruslan and Manning, Christopher D.},
  booktitle = {Proceedings of the 2018 Conference on Empirical Methods in Natural Language Processing (EMNLP)},
  year      = {2018},
  pages     = {2369--2380},
  doi       = {10.18653/v1/D18-1259}
}

@inproceedings{mallen2023popqa,
  title     = {When Not to Trust Language Models: Investigating the Effectiveness of Parametric and Non Parametric Memories},
  author    = {Mallen, Alex and Asai, Akari and Zhong, Victor and Das, Rajarshi and Khashabi, Daniel and Hajishirzi, Hannaneh},
  booktitle = {Proceedings of the 61st Annual Meeting of the Association for Computational Linguistics (ACL)},
  year      = {2023},
  pages     = {9802--9822},
  doi       = {10.18653/v1/2023.acl-long.546}
}

@article{chang2025tracedet,
  title={TraceDet: Hallucination Detection from the Decoding Trace of Diffusion Large Language Models},
  author={Chang, Shenxu and Yu, Junchi and Wang, Weixing and Chen, Yongqiang and Yu, Jialin and Torr, Philip and Gu, Jindong},
  journal={arXiv preprint arXiv:2510.01274},
  year={2025}
}

@article{ye2025dream,
  title={Dream 7B: Diffusion Large Language Models},
  author={Ye, Jiacheng and Xie, Zhihui and Zheng, Lin and Gao, Jiahui and Wu, Zirui and Jiang, Xin and Li, Zhenguo and Kong, Lingpeng},
  journal={arXiv preprint arXiv:2508.15487},
  year={2025}
}

@article{hemmat2026tdgnet,
  title={TDGNet: Hallucination Detection in Diffusion Language Models via Temporal Dynamic Graphs},
  author={Hemmat, Arshia and Torr, Philip and Chen, Yongqiang and Yu, Junchi},
  journal={arXiv preprint arXiv:2602.08048},
  year={2026}
}

@article{qian2026dynhd,
  title={{DynHD}: Hallucination Detection for Diffusion Large Language Models via Denoising Dynamics Deviation Learning},
  author={Qian, Yanyu and Tan, Yue and Liu, Yixin and Yu, Wang and Pan, Shirui},
  journal={arXiv preprint arXiv:2603.16459},
  year={2026}
}

@inproceedings{manakul2023selfcheckgpt,
  title={{SelfCheckGPT}: Zero Resource Black Box Hallucination Detection for Generative Large Language Models},
  author={Manakul, Potsawee and Liusie, Adian and Gales, Mark J. F.},
  booktitle={Proceedings of the 2023 Conference on Empirical Methods in Natural Language Processing (EMNLP)},
  year={2023},
  url={https://arxiv.org/abs/2303.08896}
}

@inproceedings{zhang2023sac3,
  title={{SAC3}: Reliable Hallucination Detection in Black Box Language Models via Semantic aware Cross check Consistency},
  author={Zhang, Jiaxin and Li, Zhuohang and Das, Kamalika and Malin, Bradley A. and Kumar, Sricharan},
  booktitle={Findings of the Association for Computational Linguistics: EMNLP 2023},
  year={2023},
  url={https://arxiv.org/abs/2311.01740}
}

@techreport{undp2025divergence,
  title       = {The Next Great Divergence: Why {AI} May Widen Inequality Between Countries},
  author      = {{UNDP}},
  institution = {United Nations Development Programme},
  year        = {2025},
  url         = {https://www.undp.org/asia-pacific/publications/next-great-divergence}
}


\clearpage
\appendix

\section{Experimental Setup and Reproducibility}
\label{app:setup}

The evaluation covers four instruction tuned models and three open domain
factual question answering datasets. Table~\ref{tab:corpus} gives the prompt
counts used for each model and dataset. The main analysis uses $K=3$ seeds
per prompt, with all decoding parameters fixed while only the sampling seed
varies.

\begin{table}[t]
  \centering
  \small
  \setlength{\tabcolsep}{4pt}
  \begin{tabular}{llrrr}
    \toprule
    \textbf{Model} & \textbf{Corpus} & \textbf{Triv.} &
    \textbf{Hot.} & \textbf{Pop.} \\
    \midrule
    LLaDA & scaled & 1000 & 800 & 1050 \\
    Dream & dream & 200 & 200 & 200 \\
    Qwen2.5 7B & AR & 400 & 400 & 400 \\
    Llama 3.1 8B & AR & 800 & 800 & 800 \\
    \bottomrule
  \end{tabular}
  \caption{Prompt counts per model and dataset. Each prompt uses $K=3$
  seeds in the main analysis.}
  \label{tab:corpus}
\end{table}

Autoregressive generations use temperature $0.8$, nucleus probability
$p=0.95$, and seeds $\{42,123,456\}$. Diffusion generations use the
models' standard denoising samplers over $128$ steps. The diffusion scripts
do not expose a scalar sampling temperature, so no temperature is reported
for LLaDA or Dream.

The answer span is extracted with a gold free first sentence heuristic.
The token level Jaccard threshold is $\tau=0.5$ unless otherwise stated.
A generation is correct when its answer contains a gold alias. A prompt is
hallucinated when none of its sampled generations contains a gold alias.
Gold information is used only for labels and scoring, never as an input
feature.

Learned detectors use five fold out of fold cross validation with
standardization local to each fold and ten shuffles. Fixed scalar signals
are evaluated directly on their prompt level scores. Bootstrap confidence
intervals use $B=2000$ prompt level resamples. Permutation tests use
$B=2000$ shuffles.

The high depth LLaDA corpus contains $50$ prompts and $18$ seeds per dataset,
for $150$ prompts in total. It is used only for robustness over seed depth.

\section{Confidence and Aggregate Agreement}
\label{app:aggregate}

Mean denoising confidence provides little separation between correct and
wrong answers on LLaDA. Table~\ref{tab:conf} reports the complete confidence
baseline.

\begin{table}[t]
  \centering
  \small
  \setlength{\tabcolsep}{3.5pt}
  \begin{tabular}{@{}lccccc@{}}
    \toprule
    & \textbf{Conf.} & \textbf{Conf.} & & \textbf{Conf.} &
    \textbf{\% high} \\
    \textbf{Dataset} & \textbf{correct} & \textbf{wrong} &
    \textbf{ECE} & \textbf{error AUC} & \textbf{wrong} \\
    \midrule
    TriviaQA & 0.951 & 0.950 & 0.52 & 0.502 & 56.4 \\
    HotpotQA & 0.955 & 0.959 & 0.72 & 0.502 & 76.7 \\
    PopQA & 0.964 & 0.965 & 0.64 & 0.497 & 67.5 \\
    \bottomrule
  \end{tabular}
  \caption{LLaDA confidence baseline. Mean confidence is similar for
  correct and wrong answers, while confidence to error AUC remains near
  chance.}
  \label{tab:conf}
\end{table}

Agreement across seeds is more informative. Hallucinated prompts disagree
more than correct prompts on all three datasets, with Mann Whitney
$p<10^{-6}$. As a single scalar feature, disagreement gives AUC
$0.61$, $0.61$, and $0.66$ on TriviaQA, HotpotQA, and PopQA respectively
(Table~\ref{tab:aggregate}).

\begin{table}[t]
  \centering
  \small
  \begin{tabular}{lcccc}
    \toprule
    & \textbf{Disag.} & \textbf{Disag.} & & \textbf{First} \\
    \textbf{Dataset} & \textbf{hall.} & \textbf{correct} &
    \textbf{AUC} & \textbf{sent. AUC} \\
    \midrule
    TriviaQA & 0.41 & 0.24 & 0.61 & 0.71 \\
    HotpotQA & 0.51 & 0.34 & 0.61 & 0.64 \\
    PopQA & 0.59 & 0.34 & 0.66 & 0.73 \\
    \bottomrule
  \end{tabular}
  \caption{Aggregate agreement across seeds on LLaDA. The first sentence
  variant uses content word disagreement in the first sentence.}
  \label{tab:aggregate}
\end{table}

\section{Full Agreement Results and Circularity}
\label{app:agreement}

Table~\ref{tab:ceiling-full} reports the complete raw agreement result for
all twelve model and dataset settings.

\begin{table*}[tb]
  \centering
  \small
  \setlength{\tabcolsep}{3pt}
  \begin{tabular}{llrrrccc}
    \toprule
    \textbf{Model} & \textbf{Dataset} & $N$ & \textbf{hall.\%} &
    \textbf{Ghost\%} & \textbf{Ghost AUC} &
    \textbf{Flick.\ AUC} & \textbf{Gap [95\% CI]} \\
    \midrule
    LLaDA & TriviaQA & 1000 & 50.0 & 77.4 & 0.506 & 0.963 &
    $0.457\,[0.429,0.485]$ \\
    LLaDA & HotpotQA & 800 & 72.2 & 69.9 & 0.479 & 0.900 &
    $0.422\,[0.390,0.458]$ \\
    LLaDA & PopQA & 1050 & 61.9 & 65.1 & 0.521 & 0.917 &
    $0.396\,[0.365,0.425]$ \\
    Dream & TriviaQA & 200 & 49.5 & 30.3 & 0.511 & 0.881 &
    $0.370\,[0.302,0.441]$ \\
    Dream & HotpotQA & 200 & 75.5 & 28.5 & 0.340 & 0.796 &
    $0.455\,[0.388,0.519]$ \\
    Dream & PopQA & 200 & 61.0 & 16.4 & 0.443 & 0.795 &
    $0.352\,[0.279,0.425]$ \\
    Qwen & TriviaQA & 400 & 36.8 & 76.9 & 0.515 & 0.948 &
    $0.432\,[0.382,0.485]$ \\
    Qwen & HotpotQA & 400 & 63.2 & 59.7 & 0.478 & 0.899 &
    $0.421\,[0.375,0.468]$ \\
    Qwen & PopQA & 400 & 54.0 & 58.8 & 0.538 & 0.905 &
    $0.367\,[0.322,0.413]$ \\
    Llama & TriviaQA & 800 & 29.1 & 66.5 & 0.546 & 0.927 &
    $0.381\,[0.344,0.418]$ \\
    Llama & HotpotQA & 800 & 71.0 & 62.5 & 0.403 & 0.837 &
    $0.434\,[0.405,0.463]$ \\
    Llama & PopQA & 800 & 49.6 & 65.0 & 0.549 & 0.905 &
    $0.356\,[0.327,0.387]$ \\
    \bottomrule
  \end{tabular}
  \caption{Raw agreement gap under disagreement score $1-\hat\pi$.
  Bootstrap confidence intervals use $B=2000$ prompt level resamples.}
  \label{tab:ceiling-full}
\end{table*}

The raw Ghost AUC ranges from $0.34$ to $0.55$, while Flickering AUC ranges
from $0.80$ to $0.96$. The raw gap ranges from $0.35$ to $0.46$ and every
gap confidence interval excludes zero. The result is partly mechanical because the disagreement signal is closely
related to the statistic used to define the regimes. Table~\ref{tab:spearman}
reports the correlation between plurality and disagreement.

\begin{table}[t]
  \centering
  \small
  \setlength{\tabcolsep}{4pt}
  \begin{tabular}{llc}
    \toprule
    \textbf{Model} & \textbf{Dataset} &
    $\rho(\hat\theta,1-\hat\pi)$ \\
    \midrule
    LLaDA & TriviaQA & $-0.996$ \\
    LLaDA & HotpotQA & $-0.991$ \\
    LLaDA & PopQA & $-0.998$ \\
    Dream & TriviaQA & $-1.000$ \\
    Dream & HotpotQA & $-0.939$ \\
    Dream & PopQA & $-0.985$ \\
    Qwen & TriviaQA & $-0.998$ \\
    Qwen & HotpotQA & $-0.993$ \\
    Qwen & PopQA & $-0.998$ \\
    Llama & TriviaQA & $-0.995$ \\
    Llama & HotpotQA & $-0.990$ \\
    Llama & PopQA & $-0.997$ \\
    \bottomrule
  \end{tabular}
  \caption{Mechanical coupling between plurality and disagreement.
  The raw agreement gap is therefore treated as descriptive rather than
  as independent evidence.}
  \label{tab:spearman}
\end{table}

\section{Decoupled Response Level Controls}
\label{app:decoupled}

The Ghost and Flickering assignments are fixed using the plurality fraction.
Two response level measurements are then evaluated without reusing the
answer span, clustering procedure, plurality statistic, or Jaccard threshold.

Lexical dispersion is the mean pairwise $1$ minus Jaccard distance over
content words in the full responses. Semantic dispersion is the mean
pairwise cosine distance between normalized all MiniLM L6 v2 embeddings
of the full responses.

\begin{table*}[tb]
  \centering
  \scriptsize
  \setlength{\tabcolsep}{2.5pt}
  \resizebox{\textwidth}{!}{%
  \begin{tabular}{ll ccc ccc ccc}
    \toprule
    & & \multicolumn{3}{c}{\textbf{Raw agreement}} &
    \multicolumn{3}{c}{\textbf{Lexical dispersion}} &
    \multicolumn{3}{c}{\textbf{Semantic dispersion}} \\
    \cmidrule(lr){3-5}
    \cmidrule(lr){6-8}
    \cmidrule(lr){9-11}
    \textbf{Model} & \textbf{Dataset}
    & G & F & $\Delta$
    & G & F & $\Delta$
    & G & F & $\Delta$ \\
    \midrule
    LLaDA & TriviaQA & .506 & .963 & .457
      & .643 & .770 & .126
      & .690 & .831 & .142 \\
    LLaDA & HotpotQA & .479 & .900 & .422
      & .605 & .824 & .219
      & .642 & .814 & .172 \\
    LLaDA & PopQA & .521 & .917 & .396
      & .746 & .880 & .134
      & .809 & .891 & .083 \\
    Dream & TriviaQA & .511 & .881 & .370
      & .605 & .835 & .230
      & .618 & .732 & .115 \\
    Dream & HotpotQA & .340 & .796 & .455
      & .547 & .789 & .242
      & .534 & .724 & .190 \\
    Dream & PopQA & .443 & .795 & .352
      & .598 & .849 & .251
      & .669 & .825 & .156 \\
    Qwen & TriviaQA & .515 & .948 & .432
      & .614 & .853 & .238
      & .627 & .888 & .261 \\
    Qwen & HotpotQA & .478 & .899 & .421
      & .574 & .760 & .186
      & .612 & .810 & .198 \\
    Qwen & PopQA & .538 & .905 & .367
      & .635 & .862 & .228
      & .690 & .833 & .144 \\
    Llama & TriviaQA & .546 & .927 & .381
      & .593 & .880 & .287
      & .628 & .813 & .184 \\
    Llama & HotpotQA & .403 & .837 & .434
      & .461 & .689 & .228
      & .539 & .627 & .088 \\
    Llama & PopQA & .549 & .905 & .356
      & .628 & .809 & .181
      & .641 & .760 & .119 \\
    \bottomrule
  \end{tabular}
  }%
  \caption{Detectability gap for frozen Ghost and Flickering regimes.
  G and F denote Ghost and Flickering AUC. All twelve lexical and all
  twelve semantic confidence intervals exclude zero.}
  \label{tab:decoupled-full}
\end{table*}

The lexical gap is positive in all twelve settings, ranging from $0.13$ to
$0.29$. The semantic gap is also positive in all twelve settings, ranging
from $0.08$ to $0.26$. The asymmetry is smaller than the raw agreement gap,
but remains positive after the mechanically coupled component is removed.

\section{Entailment Based Regime Robustness}
\label{app:entailment}

The Ghost and Flickering partition uses a gold free answer span with token level
Jaccard clustering, which is a lexical proxy for whether two samples express the
same answer. A stricter and widely used criterion instead clusters samples by
bidirectional entailment, so that two answers are grouped only when each entails
the other \citep{farquhar2024detecting}. This appendix recomputes the regimes
under that criterion and checks whether the decoupled detectability gap survives.

For each hallucinated prompt, the first sentence of every seed response is taken
as the answer, and two seeds share a cluster when a natural language inference
model assigns entailment in both directions. The model is the public
\texttt{nli-deberta-v3-base} cross encoder, and clustering is single linkage.
The plurality fraction and the Ghost cut at one half are unchanged. The decoupled
lexical and semantic dispersion signals are then scored against the frozen
entailment regimes exactly as in Appendix~\ref{app:decoupled}; under the token
level plurality partition these columns reproduce Table~\ref{tab:decoupled-full}.

\begin{table*}[tb]
  \centering
  \small
  \setlength{\tabcolsep}{5pt}
  \begin{tabular}{ll c cc cc cc}
    \toprule
    & & \textbf{Reassign.} & \multicolumn{2}{c}{\textbf{Ghost \%}} &
    \multicolumn{2}{c}{\textbf{Lexical} $\Delta$} &
    \multicolumn{2}{c}{\textbf{Semantic} $\Delta$} \\
    \cmidrule(lr){4-5}\cmidrule(lr){6-7}\cmidrule(lr){8-9}
    \textbf{Model} & \textbf{Dataset} & \textbf{\%} &
    \textbf{plur.} & \textbf{ent.} &
    \textbf{plur.} & \textbf{ent.} &
    \textbf{plur.} & \textbf{ent.} \\
    \midrule
    LLaDA & TriviaQA & 38.4 & 77.4 & 43.8 & .126 & .164 & .142 & .223 \\
    LLaDA & HotpotQA & 29.9 & 69.9 & 43.4 & .219 & .187 & .172 & .171 \\
    LLaDA & PopQA & 47.2 & 65.1 & 19.7 & .134 & .163 & .083 & .170 \\
    Dream & TriviaQA & 26.3 & 30.3 & 20.2 & .230 & .192 & .115 & .165 \\
    Dream & HotpotQA & 22.5 & 28.5 & 11.3 & .242 & .321 & .190 & .390 \\
    Dream & PopQA & 14.8 & 16.4 & 6.6 & .251 & .322 & .156 & .085 \\
    Qwen & TriviaQA & 42.2 & 76.9 & 41.5 & .238 & .166 & .261 & .158 \\
    Qwen & HotpotQA & 32.8 & 59.7 & 30.0 & .186 & .193 & .198 & .233 \\
    Qwen & PopQA & 40.7 & 58.8 & 22.7 & .228 & .246 & .144 & .204 \\
    Llama & TriviaQA & 45.1 & 66.5 & 41.2 & .287 & .242 & .184 & .288 \\
    Llama & HotpotQA & 47.4 & 62.5 & 37.0 & .228 & .212 & .088 & .276 \\
    Llama & PopQA & 39.3 & 65.0 & 56.4 & .181 & .243 & .119 & .230 \\
    \bottomrule
  \end{tabular}
  \caption{Regime robustness under bidirectional entailment clustering. Ghost
  \% is the share of hallucinations assigned to Ghost under the token level
  plurality criterion (plur.) and the entailment criterion (ent.). Lexical and
  semantic $\Delta$ are the decoupled Flickering minus Ghost AUC gaps under each
  partition. The entailment criterion reassigns a large fraction of prompts and
  lowers Ghost prevalence, yet both decoupled gaps remain positive in all twelve
  settings.}
  \label{tab:entailment}
\end{table*}

The entailment criterion is stricter than token clustering and reassigns $15$
to $47\%$ of hallucinated prompts, which lowers the Ghost prevalence in every
setting (Table~\ref{tab:entailment}). Despite this large change in the partition, the decoupled lexical gap
remains positive in all twelve settings and the semantic gap remains positive in
all twelve settings. The cross model ordering of Ghost prevalence is preserved.
The detectability asymmetry therefore does not depend on the token level
equivalence rule, although the prevalence magnitude does.

\section{Disjoint Generation Control}
\label{app:disjoint}

The response level controls freeze the regimes and change the measurement, but
the regime assignment and the dispersion score are computed from the same
sampled responses. This appendix removes that shared dependence by assigning the
regime from one set of generations and scoring detectability on a disjoint set.

The high depth LLaDA corpus provides eighteen seeds per prompt, which supports
this split. For each of fifty random partitions the eighteen seeds are divided
into an assignment half and a scoring half of nine seeds each. The hallucination
label and the Ghost or Flickering assignment use only the assignment half, while
the lexical and semantic dispersion signals use only the scoring half. The two
halves never share a generation. Table~\ref{tab:disjoint} reports the mean gap
and the fraction of splits with a positive gap.

\begin{table}[t]
  \centering
  \small
  \setlength{\tabcolsep}{4pt}
  \begin{tabular}{lcccc}
    \toprule
    & \multicolumn{2}{c}{\textbf{Lexical} $\Delta$} &
    \multicolumn{2}{c}{\textbf{Semantic} $\Delta$} \\
    \cmidrule(lr){2-3}\cmidrule(lr){4-5}
    \textbf{Dataset} & \textbf{mean} & \textbf{\% $>0$} &
    \textbf{mean} & \textbf{\% $>0$} \\
    \midrule
    TriviaQA & 0.306 & 100 & 0.288 & 100 \\
    HotpotQA & 0.135 & 100 & 0.078 & 100 \\
    PopQA & 0.065 & 96 & 0.019 & 76 \\
    \bottomrule
  \end{tabular}
  \caption{Disjoint generation control on the high depth LLaDA corpus ($50$
  prompts, $18$ seeds), averaged over $50$ random nine by nine seed splits.
  Regime assignment uses one half of the seeds and the dispersion score uses the
  disjoint other half.}
  \label{tab:disjoint}
\end{table}

Under this stricter separation the decoupled gap remains clearly positive on
TriviaQA and HotpotQA, where both the lexical and semantic gaps are positive in
every split. On PopQA the gap is weak: the lexical gap is small and the semantic
gap is close to zero, positive in $76\%$ of splits. The asymmetry therefore
persists when assignment and scoring use disjoint generations on two of the
three datasets, while PopQA, which already has the smallest response level gap,
is not robust to this separation at this sample size. Only the high depth LLaDA
corpus supports this analysis, so it is not available for the other models.

Taken together, the raw agreement gap (Table~\ref{tab:ceiling-full}), the
decoupled lexical and semantic dispersion gaps (Table~\ref{tab:decoupled-full}),
the entailment based recomputation of the regimes (Table~\ref{tab:entailment}),
and the disjoint generation split (Table~\ref{tab:disjoint}) are four
independent operationalizations of answer agreement, spanning three different
ways of deciding when two sampled answers count as the same response: token
level span clustering, bidirectional entailment, and a split that never lets
the same generation contribute to both the regime label and the detectability
score. The gap is positive under every one of them, so it is not an artifact of
any single choice of how Ghost and Flickering are defined.

\section{Robustness Over Seed Depth}
\label{app:seeds}

The high depth LLaDA corpus contains $50$ prompts and $18$ seeds per dataset.
The plurality estimate from three seeds averaged over many subsets tracks
the eighteen seed estimate closely, with Spearman correlation approximately
$0.999$ and mean absolute error approximately $0.009$. A single draw of
three seeds is noisier, with mean absolute error approximately $0.14$
(Table~\ref{tab:seed-robustness}).

\begin{table}[t]
  \centering
  \small
  \setlength{\tabcolsep}{3pt}
  \begin{tabular}{P{3.0cm}P{2.0cm}P{2.1cm}}
    \toprule
    \textbf{Analysis} & \textbf{Result} & \textbf{Interpretation} \\
    \midrule
    Three seed average vs.\ eighteen seeds
      & $\rho\approx.999$
      & closely aligned \\

    Three seed average vs.\ eighteen seeds
      & MAE $\approx.009$
      & low average error \\

    Single three seed draw vs.\ eighteen seeds
      & MAE $\approx.14$
      & individual draws are noisy \\

    Regime agreement
      & $\approx79\%$
      & most labels remain stable \\

    HotpotQA gap, $K=3$ to $18$
      & $0.244$ to $0.250$
      & gap remains positive \\
    \bottomrule
  \end{tabular}
  \caption{Seed depth robustness on the high depth LLaDA corpus.}
  \label{tab:seed-robustness}
\end{table}

Recomputing the Ghost and Flickering labels from eighteen seeds rather than
three seeds agrees on approximately $79\%$ of hallucinated prompts. The raw
agreement gap remains positive on all three datasets. On HotpotQA the gap
changes from approximately $0.244$ at three seeds to $0.250$ at eighteen
seeds.

Holding the eighteen seed partition fixed and recomputing the signal from
$K\in\{3,6,9,12,18\}$ subsets produces a roughly constant gap. This analysis
does not establish that three seeds are universally sufficient.

\section{Threshold Sensitivity}
\label{app:threshold}

The token level Jaccard threshold is swept over
$\tau\in\{0.34,0.5,0.67,1.0\}$ while the Ghost cut remains fixed at
$\hat\theta>\tfrac12$ (Table~\ref{tab:threshold}).

\begin{table}[t]
  \centering
  \small
  \setlength{\tabcolsep}{5pt}
  \begin{tabular}{lcccc}
    \toprule
    \textbf{Dataset} & $\tau=.34$ & $\tau=.50$ &
    $\tau=.67$ & $\tau=1.0$ \\
    \midrule
    TriviaQA & .456 & .457 & .452 & .452 \\
    HotpotQA & .422 & .422 & .428 & .428 \\
    PopQA & .396 & .396 & .398 & .398 \\
    \bottomrule
  \end{tabular}
  \caption{Raw agreement gap under alternative token level Jaccard
  thresholds on LLaDA.}
  \label{tab:threshold}
\end{table}

The raw gap changes by less than $0.006$ within each dataset. The lexical
decoupled gap changes by at most approximately $0.05$ per setting. Examples
of the lexical gap ranges are $[0.211,0.219]$ for LLaDA on HotpotQA,
$[0.265,0.287]$ for Llama on TriviaQA, and $[0.207,0.228]$ for Qwen on
PopQA. The qualitative conclusion does not depend on the selected threshold.

\section{Strict Control Using Diffusion Trajectories}
\label{app:trajectory}

The trajectory analysis provides a stronger independence check because it
uses no information across seeds. For each answer token, the analysis
records normalized commit timing, the fraction of low confidence before
commit, maximum and mean confidence before commit, confidence at commit,
a one step confidence jump, and a short confidence ramp.

These quantities are summarized into a $21$ dimensional prompt level vector.
Agreement, plurality, answer span clustering, and all other cross seed
consistency features are excluded. Trajectory features are available for
$83$ to $98\%$ of prompts depending on dataset. Prompts without a resolvable
trajectory are excluded before cross validation.

A single $\ell_2$ logistic detector is trained using repeated five fold
out of fold cross validation. The resulting scores are evaluated separately
on frozen Ghost and Flickering groups.

\begin{table}[t]
  \centering
  \scriptsize
  \setlength{\tabcolsep}{2pt}
  \begin{tabular}{llcccc}
    \toprule
    \textbf{Model} & \textbf{Dataset} & \textbf{Ghost} &
    \textbf{Flick.} & $\Delta$ [95\% CI] & $p$ \\
    \midrule
    LLaDA & TriviaQA & .631 & .737 & $.106\,[.033,.178]$ & .003 \\
    LLaDA & HotpotQA & .598 & .721 & $.123\,[.076,.170]$ & $<.001$ \\
    LLaDA & PopQA & .794 & .886 & $.092\,[.060,.123]$ & $<.001$ \\
    Dream & TriviaQA & .602 & .637 & $.035\,[-.081,.153]$ & .60 \\
    Dream & HotpotQA & .626 & .699 & $.072\,[-.039,.185]$ & .18 \\
    Dream & PopQA & .645 & .808 & $.163\,[.035,.291]$ & .003 \\
    \bottomrule
  \end{tabular}
  \caption{Trajectory only control. The detector uses dynamics within a
  single seed and no information across seeds.}
  \label{tab:trajectory-app}
\end{table}

On LLaDA the gap is positive on all three datasets, with confidence
intervals excluding zero and permutation $p<0.005$. On Dream the gap is
positive on all three datasets but reaches significance only on PopQA.

Ghost AUC is above chance under trajectory information, ranging from
$0.60$ to $0.79$. The result therefore does not indicate that Ghost
hallucinations are impossible to detect. It indicates that they remain
harder to detect than Flickering hallucinations.

\section{Model Dependent Regime Prevalence}
\label{app:prevalence}

The proportion of hallucinations assigned to Ghost varies substantially
across models (Table~\ref{tab:prevalence}).

\begin{table}[t]
  \centering
  \small
  \begin{tabular}{llc}
    \toprule
    \textbf{Model} & \textbf{Architecture} &
    \textbf{Ghost share of hallucinations} \\
    \midrule
    LLaDA & diffusion & $65$ to $77$ \\
    Dream & diffusion & $16$ to $30$ \\
    Qwen2.5 7B & autoregressive & $59$ to $77$ \\
    Llama 3.1 8B & autoregressive & $62$ to $67$ \\
    \bottomrule
  \end{tabular}
  \caption{Ghost prevalence across the three datasets.}
  \label{tab:prevalence}
\end{table}

Ghost is the larger share of hallucinations in LLaDA, Qwen, and Llama, but a
minority in Dream. These prevalence values are computed under the token level
plurality criterion and are sensitive to how answer equivalence is judged.
Recomputing the regimes with bidirectional entailment clustering of the sampled
answers reassigns a fraction of hallucinated prompts and reduces the Ghost
share, yet the cross model ordering is preserved and the decoupled lexical gap
remains positive in all twelve settings (Appendix~\ref{app:entailment}). The
prevalence magnitude should therefore be read as criterion dependent, whereas
the detectability asymmetry itself is robust to this choice.

\section{Matched Model Transition Analysis}
\label{app:transition}

The $600$ Dream prompts form an exact subset of the LLaDA corpus, allowing
prompt matched comparison.

The full three state transition matrix, with LLaDA as rows and Dream as
columns and states ordered as Correct, Ghost, Flickering, is

\[
\begin{bmatrix}
194 & 9 & 16\\
27 & 67 & 181\\
7 & 17 & 82
\end{bmatrix}.
\]

Among the $347$ prompts hallucinated by both models, the two state matrix is

\[
\begin{bmatrix}
67 & 181\\
17 & 82
\end{bmatrix}.
\]

\begin{table}[t]
  \centering
  \small
  \begin{tabular}{lrr}
    \toprule
    & \textbf{Dream Ghost} & \textbf{Dream Flickering} \\
    \midrule
    \textbf{LLaDA Ghost} & 67 & 181 \\
    \textbf{LLaDA Flickering} & 17 & 82 \\
    \bottomrule
  \end{tabular}
  \caption{Matched regime transitions among the $347$ prompts hallucinated
  by both models.}
  \label{tab:transition}
\end{table}

The regime changes on $57.1\%$ of jointly hallucinated prompts (Table~\ref{tab:transition}). The
Ghost to Flickering transition occurs on $181$ prompts, while the reverse
transition occurs on $17$ prompts. McNemar's test gives
$p\approx8\times10^{-36}$. Only $67$ prompts remain Ghost under both models.
The correct or hallucination label agrees on $90.2\%$ of the $600$ matched
prompts.

\section{Behavioral Audit}
\label{app:behavior}

Each model's three seeds are classified by dominant behavior among matched
hallucinated prompts (Table~\ref{tab:behavior}).

\begin{table}[t]
  \centering
  \small
  \setlength{\tabcolsep}{4pt}
  \begin{tabular}{lcc}
    \toprule
    \textbf{Behavior} & \textbf{LLaDA} & \textbf{Dream} \\
    \midrule
    Repeat one answer & 69.8\% & 19.1\% \\
    Diversify & 29.4\% & 63.7\% \\
    Fragment & not separately reported & 15.1\% \\
    Repeated abstention & at most 2.2\% & at most 2.2\% \\
    \bottomrule
  \end{tabular}
  \caption{Dominant behavior among matched hallucinated prompts.}
  \label{tab:behavior}
\end{table}

Within the Ghost to Flickering cell containing $181$ prompts, LLaDA repeats
one wrong answer on $96.7\%$ of prompts. Among these, $33.7\%$ are verbatim
repetitions and $63.0\%$ are paraphrased repetitions. Dream is diverse on
$84.0\%$ of these prompts and fragmented on $14.9\%$.

Repeated abstention is the dominant behavior on at most $2.2\%$ of prompts
for either model and occurs in only $1.1\%$ of Dream generations in this
transition cell. The transition is therefore associated with diversification
and fragmentation rather than refusal.

\section{Qualitative Examples}
\label{app:examples}

Table~\ref{tab:examples-app} gives one Ghost and one Flickering example for
each model, together with a matched model transition. In every example,
none of the three seeds contains the gold answer.

\begin{table*}[tb]
  \centering
  \footnotesize
  \setlength{\tabcolsep}{3pt}
  \begin{tabularx}{\textwidth}{@{}P{1.45cm}P{1.35cm}>{\raggedright\arraybackslash}XP{2.1cm}P{3.45cm}c@{}}
    \toprule
    \textbf{Type} & \textbf{Data} & \textbf{Question} &
    \textbf{Gold} & \textbf{Seed answers} & $\hat\theta$ \\
    \midrule
    Ghost, LLaDA & PopQA &
    What sport does Masahito Noto play? &
    football &
    baseball, baseball, and baseball &
    $1.00$ \\

    Flickering, LLaDA & PopQA &
    Who composed ``The Mission''? &
    Ennio Morricone &
    various, James Horner, and The Doors &
    $0.33$ \\

    Ghost, Dream & TriviaQA &
    In which city are the Oscar statuettes made? &
    Chicago &
    Hollywood, Los Angeles, and Los Angeles &
    $0.67$ \\

    Flickering, Dream & TriviaQA &
    Which composer wrote ``The Dam Busters March''? &
    Eric Coates &
    an English composer, John Moore, and Edward Elgar &
    $0.33$ \\

    Ghost, Qwen & HotpotQA &
    Language of the people whose principal town was Anhaica? &
    Apalachee &
    Timucua, Timucua, and Timucua &
    $1.00$ \\

    Flickering, Qwen & PopQA &
    Who was the composer of ``Hello''? &
    Masaharu Fukuyama &
    Bruno Mars, ``a covered pop song'', and ``various songs'' &
    $0.33$ \\

    Ghost, Llama & PopQA &
    Religion of St George's Cathedral? &
    Greek Orthodox &
    ``several cathedrals, cannot determine'' three times &
    $1.00$ \\

    Flickering, Llama & TriviaQA &
    Fictional school in `Please Sir'? &
    Fenn Street School &
    no verification, Fenn St.\ Secondary Modern, and
    Fenn St.\ Elementary &
    $0.33$ \\
    \midrule
    \multicolumn{6}{@{}l}{\emph{Matched LLaDA Ghost to Dream Flickering}} \\

    LLaDA to Dream & TriviaQA &
    Final, unfinished novel by Charles Dickens? &
    Edwin Drood &
    LLaDA: Bleak House three times;
    Dream: ``Tale of Our Time'', Bleak House, and Dombey &
    n/a \\
    \bottomrule
  \end{tabularx}
  \caption{Representative hallucinations across models. Ghost examples
  show repeated incorrect answers, while Flickering examples show
  divergent incorrect answers. The final row shows a matched prompt whose
  regime changes between models.}
  \label{tab:examples-app}
\end{table*}

The matched Dickens example provides a direct illustration of model
dependent regime membership. LLaDA produces Bleak House on all three seeds,
whereas Dream produces three different incorrect answers. The question is
fixed while the generation model changes the stochastic structure of the
error.

\section{Leakage and Label Robustness}
\label{app:leakage}

Several audits test whether the reported results can be explained by
information leakage or implementation artifacts.

All learned detector evaluations use five fold cross validation with
preprocessing local to each training fold. No prompt is scored by a model
trained on that prompt.

Overwriting every denoising step after a checkpoint with noise and
recomputing the features gives a maximum absolute feature difference of
$0.0$ at all $31$ checkpoints. This verifies construction from the causal
prefix.

No online or offline feature reads a gold field. Permuting the labels
collapses detection to chance. Gaussian random features remain within
$0.06$ of chance AUC. Duplicate checks pass, and corrupting the gold field
leaves features unchanged with maximum difference $0.0$.

Across these tests there is no evidence of the tested leakage or
implementation artifacts.

\subsection{Semantic Reassessment}
\label{app:semantic}

The entire subset of incorrect Ghosts is reassessed to bound the effect of
exact match grading. The reassessment covers $387$, $404$, and $423$ prompts
for TriviaQA, HotpotQA, and PopQA respectively.

A consensus answer is relabeled correct only when it passes an embedding
retrieval filter using BGE large with cosine similarity at least $0.75$ and
bidirectional DeBERTa entailment with both directions at least $0.5$.
The reassessment is automated rather than based on blinded human annotation.

\begin{table}[t]
  \centering
  \scriptsize
  \setlength{\tabcolsep}{3pt}
  \resizebox{\columnwidth}{!}{%
  \begin{tabular}{lccc}
    \toprule
    & TriviaQA & HotpotQA & PopQA \\
    \midrule
    Ghosts initially incorrect & 387 & 404 & 423 \\
    Verified correct & 4 & 2 & 0 \\
    Relabeled percentage & 1.0 & 0.5 & 0.0 \\
    Ghost percentage, original to relabeled
      & $77.4$ to $77.2$ & $69.9$ to $69.8$ & $65.1$ to $65.1$ \\
    Agreement AUC, original to relabeled
      & $.609$ to $.611$ & $.606$ to $.604$ & $.659$ to $.659$ \\
    Raw Ghost AUC, original to relabeled
      & $.506$ to $.507$ & $.479$ to $.475$ & $.521$ to $.521$ \\
    Detector recall, original to relabeled
      & $.584$ to $.580$ & $.626$ to $.627$ & $.844$ to $.844$ \\
    \bottomrule
  \end{tabular}
  }%
  \caption{Semantic reassessment of the Ghost population. Every quantity
  depending on Ghost moves by less than $0.005$.}
  \label{tab:semantic}
\end{table}

Only $4$, $2$, and $0$ Ghosts are verified correct (Table~\ref{tab:semantic}). A generous lexical
upper bound gives $11$, $7$, and $1$ possible relabelings. Every quantity
that depends on Ghost moves by less than $0.005$. The reassessment therefore
indicates that exact match grading explains only a small fraction of the
evaluated high agreement errors.

\section{Supporting Online Detector}
\label{app:detector}

The full behavioral detector is included as supporting evidence. It uses
$36$ causal features evaluated at $31$ checkpoints. The four feature blocks
are prefix information with $5$ features, consistency information with
$12$ features, diffusion commit dynamics with $9$ features, and question
priors with $10$ features.

The detector uses five fold out of fold evaluation and $\ell_2$ logistic
regression with $C=0.2$. The operating threshold is selected to control the
cumulative false positive rate at no more than $30\%$.

\begin{table}[t]
  \centering
  \small
  \begin{tabular}{lcccc}
    \toprule
    & \multicolumn{2}{c}{\textbf{AUC within group}} &
    \multicolumn{2}{c}{\textbf{Recall at FPR 30}} \\
    \cmidrule(lr){2-3}\cmidrule(lr){4-5}
    \textbf{Dataset} & \textbf{Ghost} & \textbf{Flick.} &
    \textbf{Ghost} & \textbf{Flick.} \\
    \midrule
    TriviaQA & 0.71 & 0.97 & 58 & 92 \\
    HotpotQA & 0.69 & 0.95 & 63 & 85 \\
    PopQA & 0.88 & 0.97 & 84 & 94 \\
    \bottomrule
  \end{tabular}
  \caption{Full online detector on the scaled LLaDA corpus.}
  \label{tab:detector}
\end{table}

At cumulative FPR no greater than $30\%$, the detector achieves TPR of
$66\%$, $69\%$, and $88\%$ on TriviaQA, HotpotQA, and PopQA (Table~\ref{tab:detector}). Median
detection occurs at steps $28$, $24$, and $8$ of $128$. Recall remains
higher for Flickering than Ghost on every dataset.

These results motivate warnings before generation is complete, potentially
allowing a system to request verification before users rely on an answer.
This is a candidate safeguard for resource constrained deployments, including
in the Global South, not a demonstrated low-cost solution. A small logistic
classifier does not establish low end-to-end cost: multiple generations,
feature extraction, latency, and false alarms must also be evaluated. The
reported $30\%$ false positive ceiling further requires validation against
local needs and the consequences of unnecessary warnings.

\section{Additional Diagnostic Analyses}
\label{app:diagnostics}

The following analyses are retained in compressed form because they provide
supporting evidence without carrying the central claim.

\subsection{Per Step Agreement}
\label{app:perstep}

Recomputing answer agreement at every denoising step gives little improvement
over final answer agreement on TriviaQA and HotpotQA, with AUC changing from
$0.61$ to $0.61$ on both datasets. PopQA changes from $0.66$ to $0.79$
(Table~\ref{tab:perstep}).

\begin{table}[t]
  \centering
  \small
  \begin{tabular}{lccc}
    \toprule
    \textbf{Feature set} & \textbf{Triv.} & \textbf{Hot.} & \textbf{Pop.} \\
    \midrule
    Final answer agreement & 0.61 & 0.61 & 0.66 \\
    Per step answer agreement & 0.61 & 0.61 & 0.79 \\
    Plus text overlap & 0.70 & 0.67 & 0.80 \\
    Full consistency block & 0.70 & 0.70 & 0.81 \\
    Full detector & 0.73 & 0.75 & 0.90 \\
    \bottomrule
  \end{tabular}
  \caption{Per step agreement compared with final answer agreement on
  LLaDA.}
  \label{tab:perstep}
\end{table}

The analysis establishes that temporal resolution of agreement alone does not
eliminate the detectability asymmetry. It does not establish that transient
early disagreement across seeds is absent because convergence timing across
seeds is not directly measured.

\subsection{Transfer Across Corpora}
\label{app:transfer}

Training the online detector on two datasets and evaluating it on the held
out third gives mean transfer AUC of $0.748$, compared with $0.793$ in
domain. The signal therefore transfers partly across datasets.

\subsection{Candidate Mechanism Probe}
\label{app:mechanism}

A correlational test of subject popularity as an explanation for Ghost
membership on PopQA gives AUC $0.41$. This result is below chance and does
not support popularity as an explanation for the hard regime.

Stability under stochastic perturbation, early commitment, and collapse onto
a popularity prior remain hypotheses rather than demonstrated mechanisms.
The present analyses do not establish a causal explanation for why some
hallucinations remain highly consistent.

\section{Claim and Evidence Summary}
\label{app:claims}

Table~\ref{tab:claims} summarizes the evidence hierarchy.

\begin{table*}[tb]
  \centering
  \footnotesize
  \setlength{\tabcolsep}{3pt}
  \begin{tabular}{@{}P{2.3cm}P{2.5cm}P{2.8cm}P{2.8cm}P{2.4cm}@{}}
    \toprule
    \textbf{Claim} & \textbf{Evidence} & \textbf{Establishes} &
    \textbf{Dependence status} & \textbf{Caveat} \\
    \midrule
    Raw agreement gap &
    Four models and three datasets &
    Descriptive agreement based asymmetry &
    Mechanically coupled &
    Not independent evidence \\

    Lexical asymmetry persists &
    Positive gap in all twelve settings &
    Asymmetry survives a decoupled lexical measurement &
    No exact computational reuse &
    Still a dispersion measure \\

    Semantic asymmetry persists &
    Positive gap in all twelve settings &
    Asymmetry survives a decoupled semantic measurement &
    No shared lexical computation &
    Still a dispersion measure \\

    Trajectory asymmetry &
    LLaDA significant on all three datasets; Dream positive on all three &
    Within seed dynamics contain additional information &
    No information across seeds &
    Limited to diffusion models \\

    Prevalence varies &
    Ghost share from $16\%$ to $77\%$ &
    Error population composition depends on model &
    Descriptive &
    Four models only \\

    Regime membership varies &
    $181$ versus $17$ matched transitions &
    Prompt identity alone does not determine regime &
    Descriptive matched result &
    One matched model pair \\

    Seed depth robustness &
    Three to eighteen seeds &
    Gap remains positive at greater seed depth &
    Robustness evidence &
    LLaDA only \\
    \bottomrule
  \end{tabular}
  \caption{Summary of the central evidence. The trajectory analysis is the
  only analysis in which the evaluated signal uses no information across
  seeds.}
  \label{tab:claims}
\end{table*}

\section{Statistical Procedures}
\label{app:stats}

Agreement and correctness contrasts use two sided Mann Whitney tests.
The reported values are approximately $5\times10^{-12}$ for TriviaQA,
$3\times10^{-7}$ for HotpotQA, and $1\times10^{-20}$ for PopQA.

Learned detectors use five folds with predictions held out from training
and averaged over shuffles. Bootstrap confidence intervals use $B=2000$
prompt level resamples. Each bootstrap resample redraws prompts and
recomputes both subgroup AUCs, so the gap intervals are paired.

Permutation tests use $B=2000$ shuffles of Ghost and Flickering assignments
among hallucinated prompts while detector scores remain fixed. The finite
sample convention is

\begin{equation}
p=\frac{b+1}{B+1}.
\end{equation}

The smallest reportable value is approximately $5\times10^{-4}$, which is
reported as $p<0.001$ when appropriate.

The analyses beyond the primary decoupled result are exploratory robustness
analyses and are not corrected for multiple comparisons.

\section{Aggregate AUC Decomposition}
\label{app:auc}

When the hallucination population is a mixture of Ghost and Flickering
regimes evaluated against the same correct population, aggregate AUC is a
prevalence weighted combination of the two regime specific AUCs.

Let $\pi_G$ denote the Ghost prevalence among hallucinations. Then

\begin{equation}
\begin{aligned}
\mathrm{AUC}_{\mathrm{agg}}
&=
\Pr(S_h>S_c)
+\frac{1}{2}\Pr(S_h=S_c) \\
&=
\sum_{r\in\{G,F\}}
\Pr(R_h=r\mid Y_h=1)\,
\mathrm{AUC}_r \\
&=
\pi_G\mathrm{AUC}_G
+
(1-\pi_G)\mathrm{AUC}_F .
\end{aligned}
\label{eq:auc-decomp}
\end{equation}

Here $S_h$ and $S_c$ denote detector scores for hallucinated and correct
examples, while $R_h$ denotes the regime of a hallucinated example.
Aggregate performance therefore depends jointly on regime prevalence and
detectability conditioned on regime.

Because Ghost prevalence varies substantially across models, aggregate AUC
can change as the composition of the hallucination population changes, even
when regime conditioned detector behavior is similar.

\section{Scope and Limitations}
\label{app:limitations}

The experiments cover four models, one size range, one sampling configuration
per model, and $K=3$ seeds in the main analysis. The high depth robustness
study contains $150$ LLaDA prompts and does not provide a matched seed depth
study for autoregressive models.

Correctness is determined by gold alias containment. The diffusion corpora
use token level matching, while the autoregressive corpora use a case
insensitive substring match. These procedures can misjudge paraphrases and
formatting variants.

The assignment of hallucinations to the Ghost and Flickering regimes depends on
how answer equivalence is judged. The reported partition uses a gold free
answer span with token level clustering, which can group answers that share
surface tokens without sharing meaning. Recomputing the regimes with
bidirectional entailment clustering reassigns a fraction of hallucinated prompts
and lowers the Ghost prevalence, so prevalence magnitudes are criterion
dependent. The decoupled detectability gap remains positive in every setting
under this alternative criterion (Appendix~\ref{app:entailment}), so the central
asymmetry does not depend on the specific equivalence rule.

The semantic reassessment is automated and reduces the measured effect of
possible exact match errors, but it does not constitute human adjudication.

The trajectory analysis applies only to diffusion models. Evidence is strong
for LLaDA and directional but less precise for Dream.

The Ghost and Flickering taxonomy is an operational behavioral partition.
It identifies a reproducible distinction in the evaluated settings but does
not establish that every hallucination in other architectures or domains
belongs to the same two regimes.

The Global South framing is a deployment motivation, not a regional impact
finding. We do not measure users' trust, AI literacy, or comparative exposure
to hallucination harms, nor do these benchmarks establish performance across
local languages and knowledge contexts. Future evaluation should involve
communities in defining relevant tasks and acceptable error rates, and measure
end-to-end compute, latency, and cost before claiming affordable protection.


\end{document}